\documentclass[11pt,a4paper]{article}
\usepackage{arabtex}
\usepackage{utf8}
\usepackage[hyperref]{acl2017}
\usepackage{times}
\usepackage{latexsym}
\usepackage{url}
\usepackage{graphicx}
\usepackage{graphics}
\usepackage{tabularx}
\usepackage{verbatim}
\usepackage{fancyhdr} 
\setcode{utf8}

\aclfinalcopy

\title{Acquisition of Translation Lexicons for Historically Unwritten Languages via Bridging Loanwords}

\author{Michael Bloodgood \\
  Department of Computer Science \\
  The College of New Jersey \\
  Ewing, NJ 08628 \\
  {\tt mbloodgood@tcnj.edu} \\\And
  Benjamin Strauss \\
  Computer Science and Engineering Dept. \\
  The Ohio State University \\
  Columbus, OH 43210\\
  {\tt strauss.105@osu.edu} \\}

\date{}

\begin{document}

\thispagestyle{fancy}

\maketitle

\begin{abstract}
With the advent of informal electronic communications such as social media, colloquial languages that were historically unwritten are being written for the first time
in heavily code-switched environments. We present a method for inducing portions of translation lexicons through the use of expert knowledge in these settings where there are approximately zero resources available other than a language informant, potentially not even
large amounts of monolingual data. We investigate inducing a Moroccan Darija-English translation lexicon via French loanwords bridging into English and find that a useful lexicon is induced for human-assisted
translation and statistical machine translation. 
\end{abstract}

\section{Introduction} \label{introduction}

With the explosive growth of informal electronic communications such as email, social media, web comments, etc., colloquial languages that were historically unwritten are starting to be written for the first time. For these languages, there are extremely limited (approximately zero) resources available, not even large amounts of monolingual text data or possibly not even small amounts of monolingual text data. Even when audio resources are available, difficulties arise when converting sound to text \cite{tratz2013,robinson2003}. Moreover, the text data that can be obtained often has non-standard spellings and substantial code-switching with other traditionally written languages \cite{tratz2013}. 

In this paper we present a method for the acquisition of translation lexicons via loanwords and expert knowledge that requires zero resources of the borrowing language. Many historically unwritten languages borrow from highly resourced languages. Also, it is often feasible to locate a language expert to find out how sounds in these languages would be rendered if they were to be written as many of them are beginning to be written in social media, etc. We thus expect the general method to be applicable for multiple historically unwritten languages. 
In this paper we investigate inducing a Moroccan Darija-English translation lexicon via borrowed French words. 
Moroccan Darija is an historically unwritten dialect of Arabic spoken by millions but lacking in standardization
and linguistic resources \cite{tratz2013}. Moroccan Darija is known to borrow many words from French, one of the most highly resourced languages in the world. By mapping Moroccan Darija-French borrowings to their donor French words, we can rapidly create lexical resources for portions of Moroccan Darija vocabulary for which no resources currently exist. 
For example, we could use one of many bilingual French-English dictionaries to bridge into English and create a Moroccan Darija-English translation lexicon that can be used to assist professional translation of Moroccan Darija into English and to assist with construction of Moroccan Darija-English Machine Translation (MT) systems. 

The rest of this paper is structured as follows. Section~\ref{related} summarizes related work; section~\ref{method} explains our method; section~\ref{experiments} 
discusses experimental results of applying our method to the case of building a Moroccan Darija-English translation lexicon; and section~\ref{conclusions} concludes. 

\section{Related Work} \label{related}

Translation lexicons are a core resource used for multilingual processing of languages. 
Manual creation of translation lexicons by lexicographers is time-consuming and expensive. 
There are more than 7000 languages in the world, many of which are historically unwritten \cite{lewis2015}.
For a relatively small number of these languages there are extensive resources available that have been manually
created. 
It has been noted by others \cite{mann2001,schafer2002} that languages are organized into families and that using cognates between sister
languages can help rapidly create translation lexicons for lower-resourced languages. 
For example, the methods in \cite{mann2001} are able to detect that English {\em kilograms} maps
to Portuguese {\em quilogramas} via bridge Spanish {\em kilogramos}. This general idea has
been worked on extensively in the context of cognates detection, with `cognate' typically re-defined to include loanwords as well as true cognates. The methods use monolingual data at a minimum and many signals such as orthographic similarity, phonetic similarity,
contextual similarity, temporal similarity, frequency similarity, burstiness similarity, and topic similarity  \cite{bloodgood2017,irvine2013,kondrak2003,schafer2002,mann2001}.  Inducing translations via loanwords was specifically targeted in \cite{tsvetkov2015a,tsvetkov2015b}.
While some of these methods don't require bilingual resources, with the possible exception of small bilingual seed dictionaries, they do at a minimum require monolingual text data in the languages to be modeled and sometimes have specific requirements on the monolingual text data such as having text coming from the same time period for each of the languages being modeled. For colloquial languages that were historically unwritten, but that are now starting to be written with the advent of social media and web comments, there are often extremely limited resources of any type available, not even large amounts of monolingual text data. Moreover, the written data that can be obtained often has non-standard spellings and code-switching with other traditionally written languages. Often the code-switching occurs within words whereby the base is borrowed and the affixes are not borrowed, analogous to the multi-language categories ``V" and ``N" from \cite{mericli2012}. The data available for historically unwritten languages, and especially the lack thereof, is not suitable for previously developed cognates detection methods that operate as discussed above. In the next section we present a method for translation lexicon induction via loanwords that uses expert knowledge and requires zero resources from the borrowing language other than a language informant.

\section{Method} \label{method}

Our method is to take word pronunciations from the donor language we are using and convert them to how they would be rendered in the borrowing language if they were to be borrowed. 
These are our candidate loanwords. 
There are three possible cases for a given generated candidate loanword string:
\begin{description}
  \item[true match] string occurs in borrowing language and is a loanword from the donor language;
  \item[false match] string occurs in borrowing language by coincidence but it's not a loanword from the donor language;
  \item[no match] string does not occur in borrowing language.
\end{description}

For the case of inducing a Moroccan Darija-English translation lexicon via French we start with a French-English bilingual dictionary and take all the French pronunciations in IPA (International Phonetic Alphabet)\footnote{\url{https://en.wikipedia.org/wiki/International_Phonetic_Alphabet}} and convert them to how they would be rendered in Arabic script. For this we created a multiple-step transliteration process: 
\begin{description}
  \item [Step 1] Break pronunciation into syllables.
  \item [Step 2] Convert each IPA syllable to a string in modified Buckwalter transliteration\footnote{The modified version of Buckwalter transliteration, \url{https://en.wikipedia.org/wiki/Buckwalter_transliteration}, replaces special characters such as $<$ and $>$ with alphanumeric characters so that the transliterations are safe for use with other standards such as XML (Extensible Markup Language). For more information see \cite{habash2010}.}, which supports a one-to-one mapping to Arabic script.
  \item [Step 3] Convert each syllable's string in modified Buckwalter transliteration to Arabic script.
  \item [Step 4] Merge the resulting Arabic script strings for each syllable to generate a candidate loanword string.
\end{description}

For syllabification, for many word pronunciations the syllables are already marked in the IPA by the `.' character; if syllables are not already marked in 
the IPA, we run a simple syllabifier to complete step 1. 
For step 2, we asked a language expert to give us a sequence of rules to convert a syllable's pronunciation to modified Buckwalter transliteration. 
This is itself a multi-step process (see next paragraph for details). In step 3, we simply do the one-to-one conversion and obtain Arabic script for each syllable. 
In step 4, we merge the Arabic script for each syllable and get the generated candidate loanword string. 

The multi-step process that takes place in step 2 of the process is:
\begin{description}
  \item [Step 2.1] Make minor vowel adjustments in certain contexts, e.g., when `a' is between two consonants it is changed to `A'. 
  \item [Step 2.2] Perform bulk of conversion by using table of mappings from IPA characters to modified Buckwalter characters such as `a'$\rightarrow$`a',`k'$\rightarrow$`k', `y:'$\rightarrow$`iy', etc. that were supplied by a language expert.
  \item [Step 2.3] Perform miscellaneous modifications to finalize the modified Buckwalter strings, e.g., if a syllable ends in `a', then append an `A' to that syllable.
\end{description}

\begin{figure}[t]
    \includegraphics[width=\columnwidth]{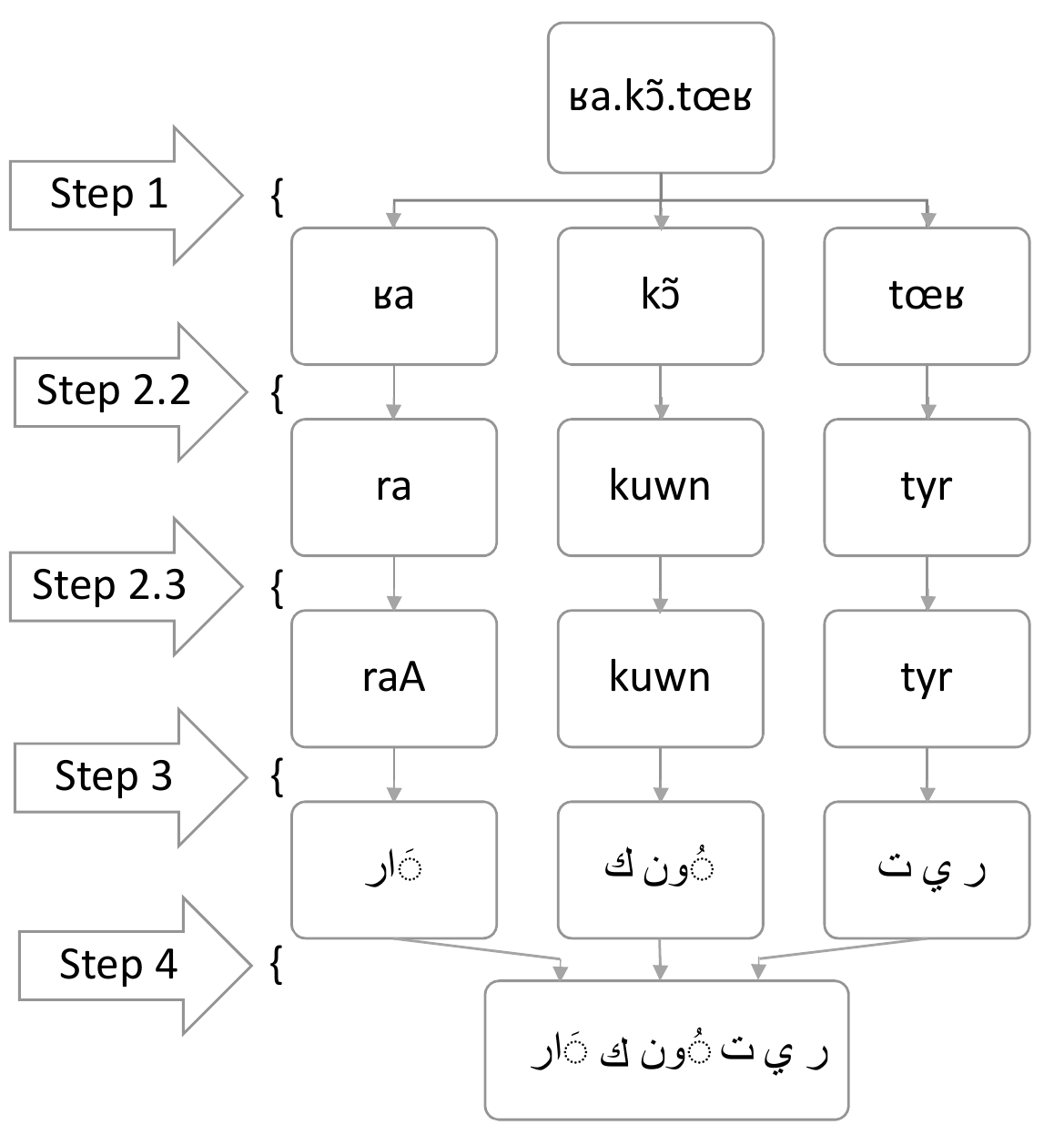}
    \caption{Example of French to Arabic Process for the French word {\em raconteur}. As discussed in the main text, step 2.1 doesn't apply to this example so it is omitted from the diagram to conserve space. Note that in the final step the word is in order of Unicode codepoints.
    Then application software that is capable of processing Arabic will render that as a proper Arabic string in right-to-left order with proper
    character joining adjustments as \<راكونتير>} \label{f:exampleConversion}
\end{figure}

The entire conversion process is illustrated in Figure~\ref{f:exampleConversion} for the French word {\em raconteur}. 
At the top of the Figure is the IPA from the French dictionary entry with syllables marked. 
At the next level, step 1 (syllabification) has been completed. 
Step 2.1 doesn't apply to any of the syllables in this word since there are no minor vowel adjustments that are applicable for this word so at the next level 
each syllable is shown after step 2.2 has been completed. 
The next level shows the syllables after step 2.3 has been completed. 
The next level shows after step 3 has been completed and then at the end the strings are merged to form the candidate loanword. 

\section{Experiments and Discussion} \label{experiments}

In our experiments we extracted a French-English bilingual dictionary using the freely available English Wiktionary dump 20131101 downloaded from 
\url{http://dumps.wikimedia.org/enwiktionary}.
From this dump we extracted all the French words, their pronunciations, and their English definitions.
Using the process described in section~\ref{method} to convert each of the French pronunciations into Arabic script yielded 8277 unique loanword candidate strings. 

The data used for testing consists of a million lines of user comments crawled from the Moroccan news website \url{http://www.hespress.com}.
The crawled user comments contain Moroccan Darija in heavily code-switched environments. While this makes for a challenging setting, it is a realistic representation of the types of environments in which historically unwritten languages are being written for the first time. The data we used is consistent with well-known code-switching among Arabic speakers, extending spoken discourse into formal writing \cite{bentahila1983,redouane2005}.
The total number of tokens in our Hespress corpus is 18,781,041. We found that 1150 of our 8277 loanword candidates appear in our Hespress corpus. Moreover, more than a million (1169087) loanword candidate instances appear in the corpus. Recall that a match could be a true match that really is a French loanword or a false match that just happens to coincidentally have string equality with words in the borrowing language, but is not a French loanword. False matches are particularly likely to occur for very short words. Accordingly, we filter out candidates that are of length less than four characters. This leaves us with 838 candidates appearing in the corpus and 217616 candidate instances in the corpus. To get an idea of what percentage of our matches are true matches versus false matches, we conducted an annotation exercise with two native Moroccan Darija speakers who also knew at least intermediate French. We pulled a random sample\footnote{We removed 15 Arabic stopwords from our candidate list before pulling the random sample.} of 1185 candidate instances from our corpus and asked each annotator to mark each instance as either:
\begin{description}
  \item[A] if the instance is originally from Arabic,
  \item[F] if the instance is originally from French, or 
  \item[U] if they were not sure.
\end{description}

The results are shown in Table~\ref{t:annotationResults}. There are a substantial number of French loanwords that are found. Some examples of translations successfully induced by our method are:
\begin{description}
\item[omelette] \<اومليت>; and
\item[bourgeoisie] \<بورجوازي>.
\end{description}

\begin{table}
\begin{center}
\begin{tabular}{ccccc} 
Annotator & Arabic & Unknown & French & Total \\ \hline
A & 907 & 88  & 190 & 1185 \\
B & 812 & 174 & 199 & 1185 \\
\end{tabular}
\end{center}
\caption {Number of word instances annotated.}  \label{t:annotationResults}
\end{table}

We hypothesize that our method can help improve machine translation (MT) of historically unwritten dialects with nearly zero resources. 
To test this hypothesis, we ran an MT experiment as follows.

First we selected a random set of sentences from the Hespress corpus that each contained at least one candidate instance and had an 
MSA/Moroccan Darija/English trilingual translator translate them into English. In total, 273 sentences were translated. This served as our test set. 
We trained a baseline MT system using all GALE MSA-English parallel corpora available from the Linguistic Data Consortium (LDC) from 2007 to 2013.\footnote{The LDC catalog numbers for the corpora we used are: LDC2008T09, LDC2007T24, LDC2008T02, LDC2009T09, LDC2009T03, LDC2012T14, LDC2012T06, LDC2012T17, LDC2012T18, LDC2013T01, and LDC2013T14.}

We trained the system using Moses 3.0 with default parameters. 
This baseline system achieves BLEU score of 7.48 on our difficult test set of code-switched Moroccan Darija and MSA.

We trained a second system using the parallel corpora with our induced Moroccan Darija-English translation lexicon appended to the end of the training data. This time the BLEU score increased to 8.11, a gain of .63 BLEU points. 

\section{Conclusions} \label{conclusions}

With the explosive growth of informal textual electronic communications such as social media, web comments, etc., many colloquial everyday languages that were historically unwritten are now 
being written for the first time often in heavily code-switched text with traditionally written languages. The new written versions of these languages pose significant 
challenges for multilingual processing technology due to Out-Of-Vocabulary (OOV) challenges. Yet it is relatively common that these historically unwritten languages borrow significant amounts of vocabulary from relatively well resourced written languages.
We presented a method for translation lexicon induction via loanwords for alleviating the OOV challenges in these settings where the borrowing language has extremely limited 
amounts of resources available, in many cases not even substantial amounts of monolingual data that is typically exploited by previous cognates and loanword detection methods 
to induce translation lexicons. This paper demonstrates induction of a Moroccan Darija-English translation lexicon via bridging French loanwords using the method and in MT experiments, the addition of the induced Moroccan Darija-English lexicon increased system performance by .63 BLEU points. 

\section*{Acknowledgments}
We would like to thank Tim Buckwalter for his support and for providing us with the initial mapping of IPA syllables to their corresponding Arabic orthographies as well as the contextual adjustment rules that we used in our experiments.
\bibliography{paper}
\bibliographystyle{acl_natbib}

\end{document}